\documentclass[opinion, inline]{sigirforum}

\usepackage[dvipsnames]{xcolor}
\DeclareMathOperator*{\argmax}{arg\,max}

\def\pubissue{Working draft.}
\begin{document}
% USE TITLE CASE FOR THE TITLE
\title{A Common Misassumption in Online Experiments with Machine Learning Models}

\authors{
\author[jeunen@sharechat.co]{Olivier Jeunen}{ShareChat}{Edinburgh, UK}
}

\maketitle 
\begin{abstract}
Online experiments such as Randomised Controlled Trials (RCTs) or A/B-tests are the bread and butter of modern platforms on the web.
They are conducted continuously to allow platforms to estimate the causal effect of replacing system variant ``A'' with variant ``B'', on some metric of interest.
These \emph{variants} can differ in many aspects.
In this paper, we focus on the common use-case where they correspond to machine learning models.
The online experiment then serves as the final arbiter to decide which model is superior, and should thus be shipped.\looseness=-1

The statistical literature on causal effect estimation from RCTs has a substantial history, which contributes deservedly to the level of trust researchers and practitioners have in this ``gold standard'' of evaluation practices.
Nevertheless, in the particular case of machine learning experiments, we remark that certain critical issues remain.
Specifically, the assumptions that are required to ascertain that A/B-tests yield unbiased estimates of the causal effect, are seldom met in practical applications.
We argue that, because variants typically learn using pooled data, a lack of \emph{model interference} cannot be guaranteed.
This undermines the conclusions we can draw from online experiments with machine learning models.
We discuss the implications this has for practitioners, and for the research literature.
\end{abstract}

\section*{Randomised Controlled Trials and their Assumptions}
Randomised experiments have existed in the scientific literature for close to 140 years, first introduced in psychology~\citep{peirce1884small}.
Since then, they have been a popular topic of study in the statistical literature ---a feat often ascribed to the seminal works of~\citet{fisher1925statistical,fisher1936design}--- and are generally well-understood~\citep{imbens2015causal}.
Randomised Controlled Trials (RCTs) form the theoretical basis for the online experiments that modern web platforms run continuously~\citep{Gupta2019}, colloquially known as A/B-tests~\citep{kohavi2020trustworthy}.

Generally speaking, RCTs deal with \emph{treatments} being applied to \emph{units}, leading to certain \emph{outcomes}~\citep{rubin1974estimating}.
Typical examples from the early literature revolve around agricultural applications, where we have types of fertiliser we can apply to plots of land, which has an effect on crop yield.
In an RCT, we randomly assign units to treatment/control, and as a result, the average measured outcomes for units under control $C$  and treatment $T$ give a finite-sample estimate of the causal effect that the treatment has on the outcome, the Average Treatment Effect (ATE):
\begin{equation}
    {\rm ATE}(C \to T; Y) = \bar{Y}(T) - \bar{Y}(C).
\end{equation}
Under seemingly reasonable and light assumptions, this estimate is \emph{consistent} and \emph{unbiased}.
That is, given infinite samples, the expectation of the estimate converges to the true average causal effect of applying the treatment instead of the control~\citep{imbens2015causal}.

The assumption that is needed to support this claim, is called the Stable Unit Treatment Value Assumption (SUTVA).
The SUTVA implies that outcomes of units are independent of the outcomes of other units under different treatments.
An agricultural example of when things can go wrong is given by \citet{rubin1974estimating}: when ``\emph{the plots are in such close proximity that following rainfall plot $i$ receives fertilizer [sic] from adjacent plots.}''
In many classical cases, it is easily argued that this assumption holds due to experimental design choices and proper randomisation.
In A/B-tests, it is often assumed that there are no ``network effects'' or ``spillovers'' among units~\citep[\S 10]{Gupta2019}.
When spillovers are known to exist, alternative experimental design frameworks have been proposed in the literature.
These are typically proposed because of some shared resource among treatments (e.g. campaign budgets in online advertising~\citep{Liu2021}), or because multiple populations interplay (e.g. buyers and sellers interfering~\citep{Bajari2021}).

\citeauthor{kohavi2020trustworthy} provide some related examples of problematic cases where interference violates the SUTVA, either through direct or indirect connections~\citep[Ch. 22]{kohavi2020trustworthy}.

In this article, we focus on the simplest possible setting, devoid of network effects, budgets, or multiple populations, and argue that the SUTVA is still violated in the large majority of online experiments that typical web platforms run on a daily basis to compare machine learning models.

\section*{Online Experiments with Machine Learning Models}
Suppose we operate a platform on the web with a deployed recommender system.
To make things concrete, we have a typical contextual bandit setup where \emph{units} correspond to contexts $x \in \mathcal{X}$, in which we recommend items $a \in \mathcal{A}$ (which we refer to as \emph{actions}).
A recommendation policy maps contexts to actions, potentially stochastically:
\begin{equation}
    \pi: \mathcal{X} \to \Delta^{|\mathcal{A}|}.
\end{equation}
As is common practice, these policies are informed by user-item interactions we have collected on the platform, and they aim to optimise some metric of interest: the outcome $Y$.
The recommendation policy itself is the \emph{treatment}.
We wish to deploy a recommendation policy, but have multiple contenders.
To reliably estimate the performance that we would obtain from deploying a given recommendation policy to the entire user population, we run an A/B-test.
That is, for every new context that the system is presented with, we randomly allocate it to either $\pi_{C}$ or $\pi_{T}$.
By assuming that users are allocated consistently, or simply that all contexts are independent, we ensure that no \emph{traditional} network effects among units exist in this setup.

Nevertheless, we argue that interference \emph{does} occur when information about logged context-action-outcome triplets under policy $\pi_{C}$ is used to inform the recommendation policy $\pi_{T}$ (or vice versa).
Furthermore, this erratic experimental design is pervasive throughout the industry, either because practitioners and researchers are unaware that they are violating the SUTVA, or because they choose to ignore it (implicitly assuming the effects will be negligible).
To the best of our knowledge, this assumption has not been validated in the research literature, and should not be accepted at face value until these effects have been studied more rigorously.
In any case, when A/B-tests are often seen as the \emph{gold standard} of evaluation methods and several research papers rely on them to provide empirical validation of newly proposed methods, we believe that it is important for practitioners, authors and reviewers to be aware of this issue, and to deal with it explicitly when evaluating conclusions that are drawn from online experiments.

We highlight two common scenarios where this type of \emph{model} interference occurs: (1) when policies learn from pooled data collected under different policies (i.e. shared training data), or (2) when features used to make decisions are informed by data collected under different policies (i.e. shared behavioural features).
The first case is easy to detect, and already widespread.
The second case can be more subtle, when e.g. embeddings are used to represent contexts and actions, and they are updated throughout the experiment (even if the recommendation policies themselves are not necessarily updated).
In both these cases, a model's output at a certain time $t$ will not be independent of other policies in place, which violates the SUTVA explicitly: $\pi_{T}^{t}(x) \not\!\perp\!\!\!\perp  \pi_{C}^{t-1}(x^{\prime})$.

Naturally, this insight does not automatically dismiss all experiments that are run with similar configurations, and it is by no means our intention to undermine the credibility of previously published work.
It does, however, implore us to take a critical look at online experiments presented in the research literature, and we hope to incentivise authors to be explicit and clear when describing their experimental setup.
Indeed, A/B-tests on large-scale platforms are inherently non-reproducible for the majority of the academic audience, and we believe that some additional scrutiny may be warranted in light of the arguments presented in this paper.

In what follows, we provide empirical evidence on synthetic data to highlight how a SUTVA violation can be problematic, and how it invalidates conclusions drawn from online experiments.

\section*{Empirical Evidence of Model Interference with Pooled Data}
To empirically validate our insight that \emph{model interference} violates the SUTVA when learning from pooled data, we provide an intuitive example and experimental proof on synthetic data.\footnote{A Python notebook reproducing the synthetic data and figures presented in this section has been made available at \url{https://github.com/olivierjeunen/pooled-data-experiments}.}
%as well as real-world data for a top-$N$ recommendation setup.
%\subsection*{Synthetic Example with Multi-Armed Bandits}

In line with the setup we laid out above, we consider varying multi-armed bandit methods to learn a recommendation policy over time.
We assume there are $|\mathcal{A}|=N$ items to recommend with binary rewards: $r \sim {\rm Bernoulli}(\rho_{a})$.
Naturally, $\rho_{a}$ is unknown to the learning methods.
We simulate it to be uniformly spaced over a $[0.05, 0.15]$ interval, to represent ``clicks'' on items.

We consider Bayesian modelling methods with a ${\rm Beta}(2,10)$ prior over $\rho_{a}$ --- emulating reasonable knowledge of the system (i.e. $\widehat{\rho}(a)=0.10$ in absence of data).
For every action $a \in \mathcal{A}$, we log the reward $r$.
For convenience of notation, we will denote with $s(a)$ the number of clicks (successes) we observed for action $a$.
Analogously, $f(a)$ denotes the number of non-clicks (failures).
The data aggregation functions $s$ and $f$ can either be computed based on all available data (\emph{pooled}), or only on data collected by the learning method itself (\emph{siloed}).
We consider the following classical multi-armed bandit methods, where ties are broken randomly:

\begin{description}
    \item[MLE-Greedy] acts according to the Maximum Likelihood Estimate: $\argmax_{a \in \mathcal{A}} \frac{s(a)}{s(a)+f(a)}$,
    \item[MAP-Greedy] acts according to the Maximum A Posteriori estimate: $\argmax_{a \in \mathcal{A}} \frac{2+s(a)}{12+s(a)+f(a)}$,
    \item[$\epsilon$-Greedy] acts uniformly at random with probability $\epsilon$, and according to MAP-Greedy otherwise,
    \item[Thompson] follows a classical Thompson Sampling approach \citep[Alg. 2]{Chapelle2011},
    \item[UCB] acts according to the $q\textsuperscript{th}$ quantile of the posterior distribution~\citep{Kaufmann2012}.
\end{description}

We simulate $2\times 10^{6}$ successive rounds where every method is presented with an opportunity to act and learn from the resulting observed reward, with $N=11$ actions.
To reduce the effects of statistical variation, this process is repeated from the onset 20 times, and we report 95\% confidence intervals for the average cumulative regret.
Note that this setup is overly simplified to represent any real-world system (2--10 million training samples for only 11 actions and no contextual dependence), and it is not our goal to simulate real-world behaviour as accurately as possible.
Nevertheless, it provides a simple, intuitive and reproducible example of the consequences that occur due to a SUTVA violation in online experiments with machine learning models.

\begin{figure*}[t]
    \centering
    \includegraphics[width=\linewidth, trim={0cm 0.5cm 0cm 0.5cm}]{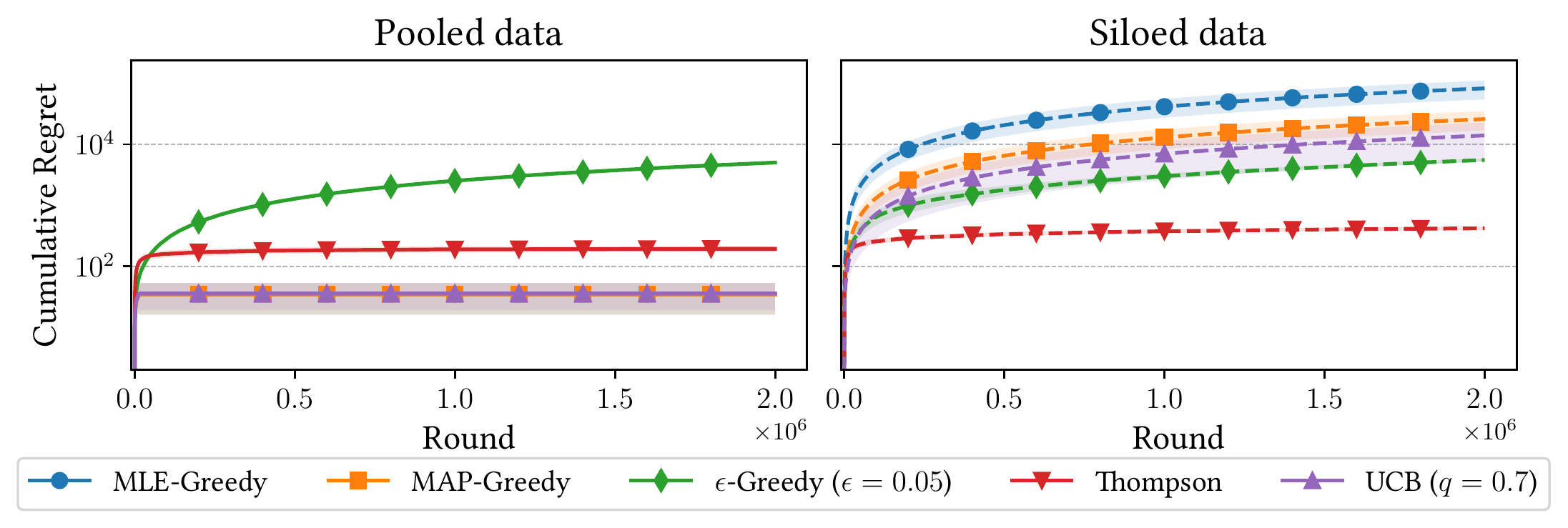}
    \caption{Cumulative regret for competing methods in an online experiment, where training data is either \emph{pooled} or \emph{siloed}.
    When siloed, the SUTVA holds, and online results are representative of what would happen if we would have shipped a method to the entire population.
    When pooled, models are faster to converge, but results are \emph{not} representative because the SUTVA is violated.
    }
    \label{fig:SUTVA_violated}
\end{figure*}

We repeat the above scenario with two configurations: \emph{pooled} data, i.e. after every round, all models are updated with the observed action-reward pairs from all variants; and \emph{siloed} data, i.e. after every round, models are only updated with the observed rewards for their own actions.
Figure~\ref{fig:SUTVA_violated} visualises 95\% confidence intervals over the cumulative regret for both configurations.
As expected, we observe disparate relative performance of competing methods under the two regimes.
Specifically, under the pooled data regime, the greedy methods (and UCB) perform exceedingly well in comparison with the other bandit algorithms ($\epsilon$-Greedy and Thompson).
This conclusion reverses when we consider the siloed data regime, where the Thompson sampling approach significantly outperforms others, and the $\epsilon$-Greedy method has gone, relative to competing methods, from the worst to the 2\textsuperscript{nd} best method (in terms of cumulative regret).

To researchers and practitioners familiar with the multi-armed bandit paradigm, it will be obvious that the \emph{pooled} data scenario cannot be reconciled with the very motivation behind these sequential decision-making methods~\citep{robbins1952some}.
Indeed, we famously need to balance \emph{exploitation} with \emph{exploration}: considering both the immediate reward an action can bring, as well as the informational value we get from executing it.
By pooling the data together, \emph{all} competing methods can reap the benefits of exploration data generated by \emph{any} of the variants.
As a result, we observe a piggybacking effect, where greedy (exploitation-heavy) methods are disproportionately favoured over exploration-heavy methods.
From the siloed data regime, however, we can clearly observe that pooled data experiments give a false representation of what would happen if the variants were shipped in isolation.
The SUTVA is violated, and this type of experimental setup is unsuitable when the goal is to provide causal estimates of the effects that different machine learning methods have on the reward that we incur from running them on the platform.

As such, we are convinced that online experiments where competing machine learning models learn from pooled data, cannot be guaranteed to provide unbiased estimates of true causal effects.
Nevertheless, we argue that this style of experiment is widespread in industry.
Often, the differences between variants might not be as outspoken as in the intuitive example laid out above, where we have exploitation-heavy methods piggybacking off of exploration-heavy alternatives.
In such cases, a SUTVA violation might be not be as detrimental to the truthfulness of the online experiment as it is in our synthetic example.
Nevertheless, the SUTVA is violated, and we have reasons to be critical about conclusions that are drawn from it.
This holds not only for the platforms that are running the experiments themselves, but also for the scientific community.
Indeed, a SUTVA violation is crucial information to have for practitioners and researchers (readers and reviewers) that wish to assess the scientific contributions of such experiments.

\section*{Why is the pooled data setup dominant in practice?}
To consider the reasons why the pooled data setup occurs so often in practice, we must consider what makes it attractive.
A first argument in favour, comes from its engineering simplicity.
Indeed, interactions between users and items that are generated over time on the platform can be logged in a shared database, and they need no further special treatment down the line.
That is, the training data for machine learning methods can simply be obtained through a scan of the database.

A second and related argument in favour, is that of training data size.
In the age of ``\emph{data hungry}'' deep learning methods~\citep{Marcus2018}, it is hard to make a case for ---optimistically--- halving the training data size of all models, or ---more realistically--- learning from only a fraction of the available training interactions, when we have multiple treatments being tested.
This is especially true when we consider a recently popular class of off-policy learning approaches that typically suffer from high variance in their objective functions~\citep{Vasile2020,Saito2022}.
In these use-cases, it comes natural to try and combat variance with an increased training sample size.
Whilst natural, we must be aware that it violates the SUTVA.

Finally, there are differing incentives when running a platform successfully as a business, compared to measuring experimental outcomes in a scientifically sound manner.
Often, the long-term goals of researchers and practitioners running online experiments in industry will be focused on \emph{improving} a given metric of interest (the outcome), and not necessarily on \emph{measuring} the effect of the treatment on the outcome as truthfully as possible.
Online experiments then become a means to an end, and minimising the number of false positives that they generate might become more important than having unbiased causal effect estimates.
Understandably so, there is a rich literature on avoiding pitfalls like false positives, related to statistical mis-reasoning in A/B-testing, and focusing on $p$-values and hypothesis testing~\citep{Kohavi2022}.

Nevertheless, when results from online experiments are adopted as empirical proof in the scientific literature, it is critical that we are explicit about the assumptions that underlie the methods that are used, and to which extent they are likely to be violated.

\section*{Conclusions and Outlook}
Offline experiments to validate the effectiveness of machine learning models in applications that rely heavily on user interactions, are notoriously hard to get right~\citep{Gilotte2018,Jeunen2019,Diaz2021,Deffayet2023}.
This is a well-known issue in the recommender systems research community, and contrasting results between on- and off-line experiments have been reported in the literature time and again~\citep{Beel2013,Garcin2014,Rossetti2016,Jeunen2018}.
The reasons underlying these diverging results, are often (rightfully) ascribed to problematic offline evaluation practices, in that they do not produce unbiased offline estimators of online metrics~\citep{Jeunen2021Thesis}.
Online experiments, on the other hand, are typically seen as the gold standard, owing to the elegant statistical theory that motivates their use.

In this short opinion paper, we have argued that this confidence might be misplaced in many practical A/B-testing scenarios.
We have briefly reviewed the main assumption that makes RCTs tick (i.e. the Stable Unit Treatment Value Assumption (SUTVA)), and have motivated how it is easily violated in the prevalent online experiment setup where we have competing machine learning methods that learn from a shared \emph{information} resource --- be it training samples or features.
When this information is somehow influenced by other machine learning methods evaluated in the experiment, this is a clear SUTVA violation.
It occurs when either the models themselves or the features they consume are updated with new (pooled) data throughout the experiment.
Through a classical multi-armed bandit setup, we have provided an empirical example that highlights how this practice can invalidate conclusions drawn from online experiments with similar violations.

As we have mentioned earlier, it is not the intention of this paper to undermine the credibility of existing work, or to dissuade the use of A/B-tests as an evaluation tool.
We do wish to cultivate an awareness of the assumptions that we (often implicitly) make when running online experiments, and point out that they are easily violated in typical modern use-cases.
As we have argued, there can be good reasons for adopting an experimental design where SUTVA is violated (e.g. to increase the training sample size for competing methods).
Nevertheless, in such cases, results obtained through and conclusions drawn from online experiments, should be presented with an explicit caveat.
This ensures that readers have access to the information that is necessary to properly assess their scientific value, and possible contributions to the state-of-the-art.

This paper raises more questions than it answers.
It is our hope that it can contribute to conversations about future work, to (1) qualitatively and quantitatively assess to what extent SUTVA violations affect experimental outcomes, (2) to develop tools and metrics that can identify problematic cases, and (3) to provide guidance and improved methods for problematic scenarios.

In particular, future work that studies the relation between the SUTVA violations we describe and the bias they impose on experimental outcomes, will be hugely impactful for both the academic and industrial research communities.
This can include both theoretical works that aim to prove bounds on the statistical bias of the estimator, as well as empirical works that quantify the extent to which different types of common online experiments with SUTVA violations can be trusted.

\section*{Acknowledgements}
%{\tt TODO}
The opinions voiced in this article have benefited from discussions with many colleagues and collaborators.
The author would like to explicitly thank Andr{\'e}s Ferraro, Thorsten Joachims and Lien Michiels for their feedback on earlier drafts.
% Lien Michiels, Andres Ferraro, Fernando Diaz, Thorsten Joachims, Maarten de Rijke, Harrie Oosterhuis, Guido Imbens, Harald Steck and Rishabh Mehrotra f

\bibliography{sigirforum}

\end{document}